\documentclass[sigconf]{acmart}

\def\BibTeX{{\rm B\kern-.05em{\sc i\kern-.025em b}\kern-.08emT\kern-.1667em\lower.7ex\hbox{E}\kern-.125emX}}

\copyrightyear{2019}
\acmYear{2019}
\setcopyright{acmlicensed}
\acmConference[MM '19]{Proceedings of the 27th ACM International Conference on Multimedia}{October 21--25, 2019}{Nice, France}
\acmBooktitle{Proceedings of the 27th ACM International Conference on Multimedia (MM '19), October 21--25, 2019, Nice, France}
\acmPrice{15.00}
\acmDOI{10.1145/3343031.3351041}
\acmISBN{978-1-4503-6889-6/19/10}

\acmSubmissionID{1000}

\usepackage{balance}

\begin{document}
\fancyhead{}

%
\title{Adversarial Colorization Of Icons Based On Structure And \\ Color Conditions}

%
\author{Tsai-Ho Sun}
\affiliation{%
  \institution{National Chiao Tung University}
  \city{Hsinchu}
  \country{Taiwan}
}
\email{locha0519@gmail.com}

\author{Chien-Hsun Lai}
\affiliation{%
  \institution{National Chiao Tung University}
      \city{Hsinchu}
  \country{Taiwan}
}
\email{jxcode.tw@gmail.com}

\author{Sai-Keung Wong}
\affiliation{%
  \institution{National Chiao Tung University}
  \city{Hsinchu}
  \country{Taiwan}
}
\email{cswingo@cs.nctu.edu.tw}

\author{Yu-Shuen Wang}
\affiliation{%
  \institution{National Chiao Tung University}
  \city{Hsinchu}
  \country{Taiwan}
}
\email{yushuen@cs.nctu.edu.tw}






%

%
\begin{abstract}
We present a system to help designers create icons that are widely used in banners, signboards, billboards, homepages, and mobile apps. Designers are tasked with drawing contours, whereas our system colorizes contours in different styles. This goal is achieved by training a dual conditional generative adversarial network (GAN) on our collected icon dataset. One condition requires the generated image and the drawn contour to possess a similar contour, while the other anticipates the image and the referenced icon to be similar in color style. Accordingly, the generator takes a contour image and a man-made icon image to colorize the contour, and then the discriminators determine whether the result fulfills the two conditions. The trained network is able to colorize icons demanded by designers and greatly reduces their workload. For the evaluation, we compared our dual conditional GAN to several state-of-the-art techniques. Experiment results demonstrate that our network is  over the previous networks. Finally, we will provide the source code, icon dataset, and trained network for public use.
\end{abstract}

%
%


%

%

%
\maketitle

\section{Introduction}
Nowadays, icons are widely utilized in banners, signboards, billboards, homepages, and mobile apps. Effective icons are usually simple but distinguishable, so that users can quickly receive the intended information when seeing them at a small size or a long distance. Considering aesthetics and practical issues, designing an eye-catching icon is challenging. Designers have to carefully consider not only shapes and structures, but also colors, when they create icons for their customers. Moreover, icons are not self existent. When they appear on a signboard or a website with letters and backgrounds, the styles of these different components should be consistent, which makes icon design more challenging. The difficulties motivate us to build a system that can reduce designers' workload. Specifically, designers draw the contour of an icon, while our system is in charge of colorization.

Generative adversarial networks (GANs) \cite{Goodfellow+:2014:GANs} have been proven to be able to generate realistic images in many applications \cite{
Cao+:2017:UnsupervisedDiverse,%
Ledig+:2017:Photo,%
li+:2016:precomputed,%
mathieu:2015:deep,%
WangGupta:2016:Generative,
lample2017fader}, and could constitute a solution to help designers colorize icons. Specifically, a network takes a contour image drawn by the designers as input and then outputs the colorized icon image.  Similar ideas have been adopted to colorize black-and-white Manga characters \cite{Ci+:MM:2018, furusawa2017comicolorization, hensman2017cGAN_Manga, zhang2017style} and achieved great success. To control the colorization process, additional inputs, such as stroke colors and style images, are fed into the network as well. The features extracted from both contour and style images will be fused and used for the colorization. However, training the above-mentioned networks to achieve icon colorization is inadequate because icons exhibit diverse styles and structures. While the discriminator is not strong enough to recognize man-made and machine-generated icons, its guidance to train the generator is inappropriate.


\begin{figure*}
    \centering
    \includegraphics[width = \linewidth]{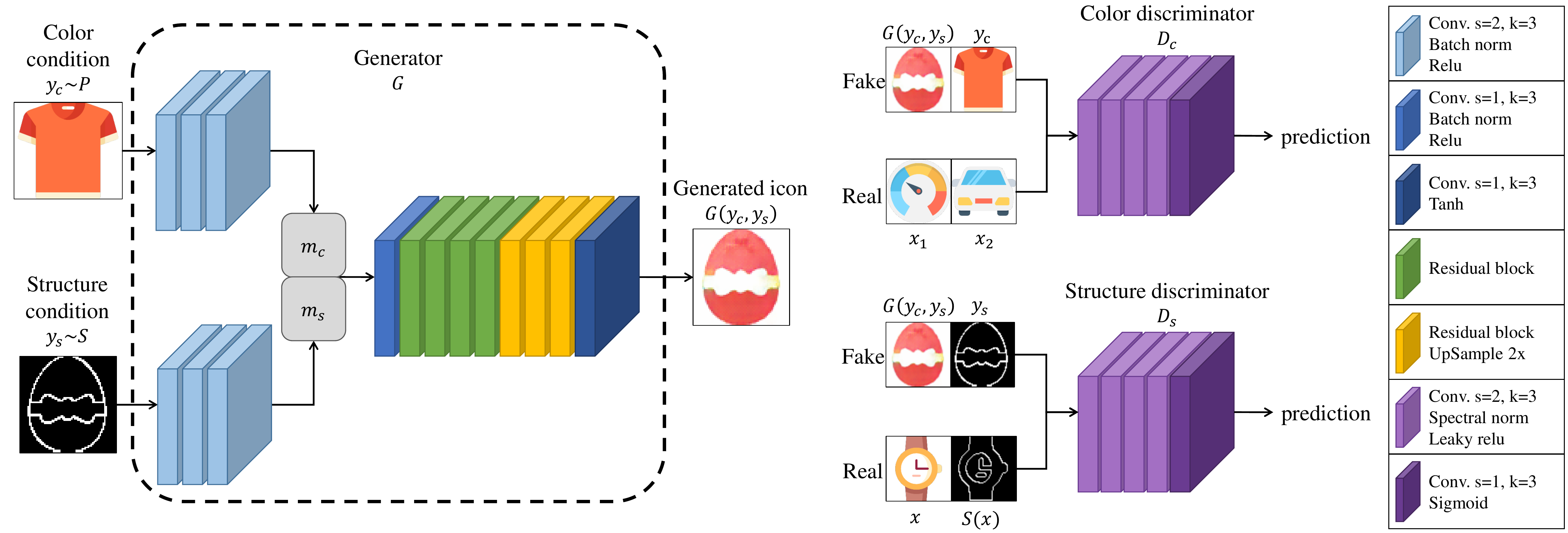}
    \caption{Our dual conditional generative adversarial network. The details of layers in different colors are on the right. $k$ and $s$ are the kernel size and the stride, respectively}
    \label{fig:network}
\end{figure*}

Observing that an icon can be well defined by color and structure conditions, we present a dual conditional GAN (Figure~\ref{fig:network}) to colorize icons. Rather than training a discriminator to recognize whether an icon is man-made or machine-generated, we train two discriminators to determine whether paired images are similar in structure and color style, respectively. In this way, the task assigned to each discriminator is simple and easy to accomplish. To specify the icon structure, we let users draw contours on a graphical interface. To condition the color style, a man-made icon image is selected. The two images are fed into our dual conditional GAN for icon colorization. Because it is not intuitive to apply a man-made icon to specify the color condition, in practice, we let users select a style label when using our system to create icons. Then the system randomly selects man-made icons that match the style \cite{Krause:1992:ColorIndex,Kobayashi:1992:color} and feeds the icons to the network for colorization.





To evaluate the performance of our icon colorization method, we tested the system on several examples with diverse structures and color styles. Figures \ref{fig:results},
\ref{fig:step_by_step}, and \ref{fig:comparison}, and our accompanying video present the results. In addition, we compared our dual conditional GAN to state-of-the-art techniques, including iGAN \cite{zhu2016generative}, CycleGAN \cite{Zhu+:2017:UnpairedImgImg}, conditional image-to-image translation \cite{Yi:2017:DualGAN}, ComiColorization \cite{furusawa2017comicolorization}, MUNIT \cite{huang2018multimodal} and Anime \cite{zhang2017style}. Experiment results demonstrate the effectiveness of our technique. 



\section{Related Work}
%
%
%
\textbf{Icon Design. }
Although icons are widely used nowadays, creating visually appealing icons is not easy because many aspects, such as context, color, and structure, should be considered \cite{Horton:1994:icon,Hawthorn:IwC:2000,Heim:2007:Resonant}. Extensive studies have conducted on issues about the message quality, metaphor, and styling of icons \cite{Huang+:2002:IFIE}, and the effects of icon spacing and size \cite{Lindberg:2003:Display}. Among the above-mentioned aspects, color is visually essential to make icons attractive, legible and viewer-friendly \cite{Lim:2010:VIC}. The optimal choice of color combinations and icon shapes can convey information both clearly and pleasantly. 




\textbf{Generative Adversarial Networks. }
GAN was first presented by Goodfellow \emph{et al.} \cite{Goodfellow+:2014:GANs} and then widely used in realistic image generation \cite{tulyakov2017mocogan, xiong2018learning, Cao+:2017:UnsupervisedDiverse, Isola+:2017:ImagetoImageTW, Ledig+:2017:Photo, li+:2016:precomputed, mathieu:2015:deep, Taigman+:2016:UnsupervisedImg, WangGupta:2016:Generative, lample2017fader}. The network typically contains a generator and one or multiple discriminators, which are trained iteratively and alternatively to surpass one another. In spite of tremendous advantages, training a GAN is challenging because of gradient vanish and stability problems. To facilitate network training, several methods, such as energy-based GANs \cite{zhao2016energy}, minibatch discrimination \cite{salimans2016improved}, Wasserstein GANs \cite{Arjovsky+:2017:WGAN, gulrajani2017improvedWGAN}, boundary equilibrium GANs \cite{berthelot2017began}, and spectral normalization \cite{Miyato+:2018:SpectralNF}, were presented.

\textbf{Conditional GANs and Domain Transfer. }
Earlier versions of GANs are not controllable because the inputs are noise latent vectors. Afterward, to control the results, additional features or conditions are fed into the network. The features can be labels \cite{mirza2014conditional,Yan+:2016:Attribute2image,lample2017fader}, images \cite{Isola+:2017:ImagetoImageTW, Liu+:2017:AutoPainter}, and sentences \cite{Bodla+:2018:SemisupervisedFF,zhang2017stackgan}. Although results generated by the conditional GANs are impressive, most of them are supervised; and they demand a large number of labels and matching image pairs. Thus, unsupervised techniques were presented to map images from one domain to another \cite{Taigman+:2016:UnsupervisedImg, Zhu+:2017:UnpairedImgImg, Yi:2017:DualGAN}. The methods retain either latent codes or reconstructions during the domain transformation to achieve the goal. To avoid the mode collapse problem that frequently occurs in GANs, Zhu \emph{et al.}~\cite{Zhu+:2017:Toward} combined the conditional variational autoencoder GAN \cite{larsen:2015:autoencoding} and the conditional latent regressor GAN \cite{donahue:2016:adversarial, dumoulin:2016:adversarially} to generate diverse and realistic results.

\textbf{Manga and Cartoon Colorization. }
In the past, coloring black-and-white Manga was achieved by considering hand crafted features, such as pattern-continuity and intensity-continuity \cite{Qu:2006:MangaColorization}. While such features are difficult to define, deep neural networks were presented to achieve the goal by learning from data automatically. By providing a line art and several guided stoke colors, the system can be used to colorize Manga \cite{Ci+:MM:2018}. In addition to stroke colors, several methods let users provide reference images for guiding results. Among them, Furusawa \emph{et al.} \cite{furusawa2017comicolorization} adopted a convolutional encoder-decoder network with an additional discriminator to colorize black-and-white Manga pages. Hensman~\emph{et al.}'s method \cite{hensman2017cGAN_Manga} required only one single reference image to train the conditional GAN \cite{Isola+:2017:ImagetoImageTW}, and used the network to colorize monochrome images. Zhang \emph{et al.} \cite{zhang2017style} fused the high level features extracted from sketch and style images to generate the color version of a sketch. They also applied two guided decoders to prevent the residual U-net from skipping high level features.

The aforementioned networks have achieved great success on colorizing Manga characters. For our icon colorization task, however, they may fail because icons are not only diverse in color style but also in structure. Furthermore, we want to achieve that boundaries of regions are formed by color difference between the regions. Because man-made icons are difficult to define, a single discriminator cannot determine whether the generated result is meaningful. To tackle this problem, we presented a dual conditional GAN that anticipates the generated results to fulfill structure and color conditions. Because the task assigned to each discriminator is simple, the network is easy to train and able to generate visually appealing results.



\section{Structure and Color of an Icon}
Given a contour image and a man-made icon, our system strives to generate a result, in which the structure is similar to the contour and the color style is similar to the referenced man-made icon. To achieve this goal, we train a conditional generative adversarial network on an icon dataset\footnote{https://www.flaticon.com/} that contains 12,575 images. These icons contain multiple colors, with a white background, but without dark border lines. We apply image processing techniques to automatically extract contours and color styles from these icons for network training. No manual labeling tasks are needed.

\subsection{Structure Condition}
We represent the structure condition by a binary contour image. White and black pixels in the image indicate edge and non-edge regions, respectively. To obtain this contour image, the Canny edge detection algorithm is adopted. Intuitively, each icon and its corresponding contour can match, and otherwise cannot.

\subsection{Color Condition}
\label{secton:colorcondition}
The color condition is specified by the referenced icon image. To determine whether two icons match in color style, we compute a 3D Lab color histogram (8 $\times$ 8 $\times$ 8) of each of them, and then measure their distance. In the pre-processing step, we apply the $K$-means ($K=500$) clustering method to merge icons if their color histograms are close to each other. Icons in the same cluster are considered similar in color style.


\section{Network Training}
We train a dual conditional GAN to colorize icons. Figure \ref{fig:network} shows the network architecture and the details of each layer. The inputs of the network are a contour image and a referenced icon image, with a resolution $64 \times 64 \times 3$. The former and the latter inputs are the structure and the color conditions, respectively. They are encoded, concatenated together, and then fed into the generator. The generator then takes the code to generate a fake icon. After that, two discriminators are used to guide the generator creating results that can fulfill the input conditions. To train the structure discriminator, the contour image is combined with the corresponding icon and the generated icon to form the real and the fake pairs, respectively; and then they are fed into the structure discriminator. To train the color discriminator, the real pair is formed by two arbitrary icons that are classified in the same cluster, whereas the fake pair is formed by the referenced and the generated icon. It is worth noting that the purpose of this discriminator is to judge whether two icons are similar in color style. The real and the fake pairs can be dissimilar in structure.



\subsection{Loss functions}
The dual conditional GAN is trained by optimizing two adversarial losses. Let $P$ and $S$ be the distributions of man-made icons and the corresponding contours, respectively. We also let $x \in P$, and $S(x)$ be the contour of $x$. To fulfill the structure condition, the loss is defined as:
\begin{align}
\nonumber L_s(G,D_s) = & \mathbb{E}_{y_c\sim P, y_s\sim S}[\log(1-D_s(G(y_c, y_s),y_s))] + \\
& \mathbb{E}_{x\sim P}[\log D_s(x,S(x))],
\end{align}
where $G(y_c, y_s)$ is the result generated according to a reference icon $y_c$ and a contour image $y_s$; and $D_s$ is a conditional discriminator that determines whether the paired images have the same structure. To fulfill the color condition, we apply a similar strategy. Let $k(x)$ be the cluster index of $x$. We present the loss as:
\begin{align}
\nonumber L_c(G,D_c) = & \mathbb{E}_{y_c\sim P, y_s\sim S}[\log(1-D_c(G(y_c, y_s),y_c))] + \\
& \mathbb{E}_{x_1, x_2\sim P, k(x_1) = k(x_2)}[\log D_c(x_1,x_2)],
\end{align}
where $D_c$ is a conditional discriminator that determines whether two images are similar in colors. Recall that we apply the $K$-means clustering to group icons that are similar in color. The real pair, $x_1$ and $x_2$, are two arbitrary icons in the same cluster. Finally, the full objective function is
\begin{align}
G^* = \arg \min_{G} \max_{D_s, D_c} \left( L_s(G,D_s) + L_c(G,D_c)\right).
\end{align}

\begin{figure}
    \centering
    \includegraphics[width = \columnwidth]{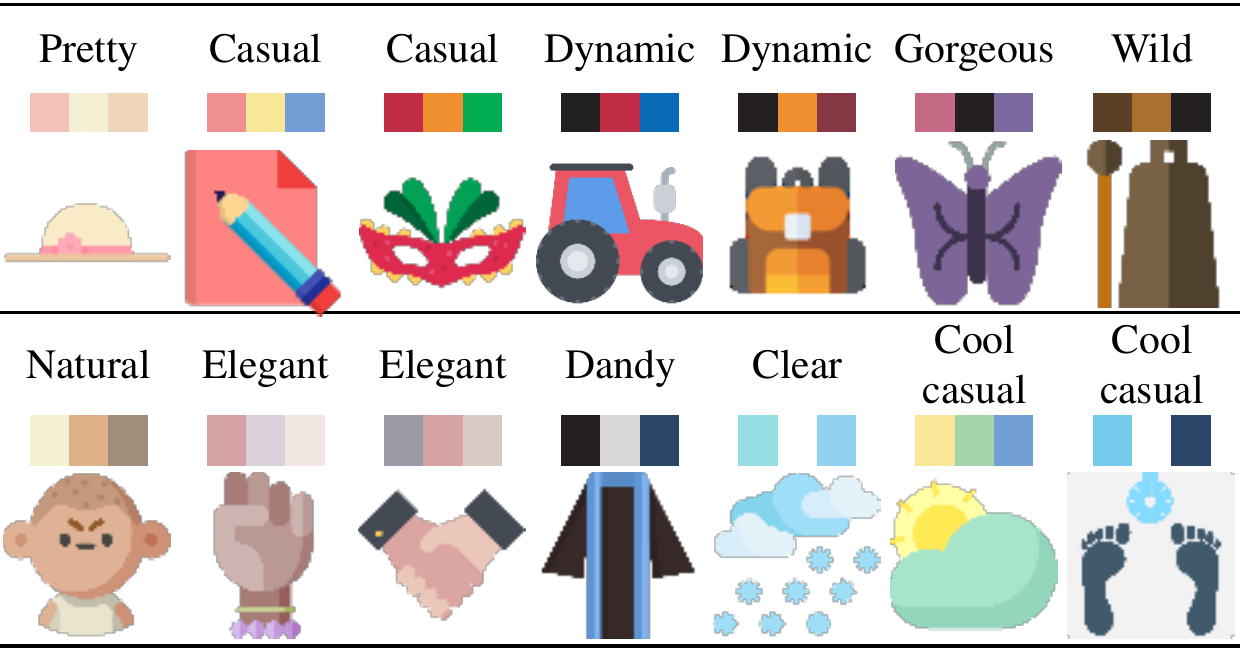}
    \caption{The semantic labels, corresponding color combinations, and the example icons.}
    \label{fig:colorCombanations.pdf}
\end{figure}

\begin{figure}
\centering
\includegraphics[width = \columnwidth]{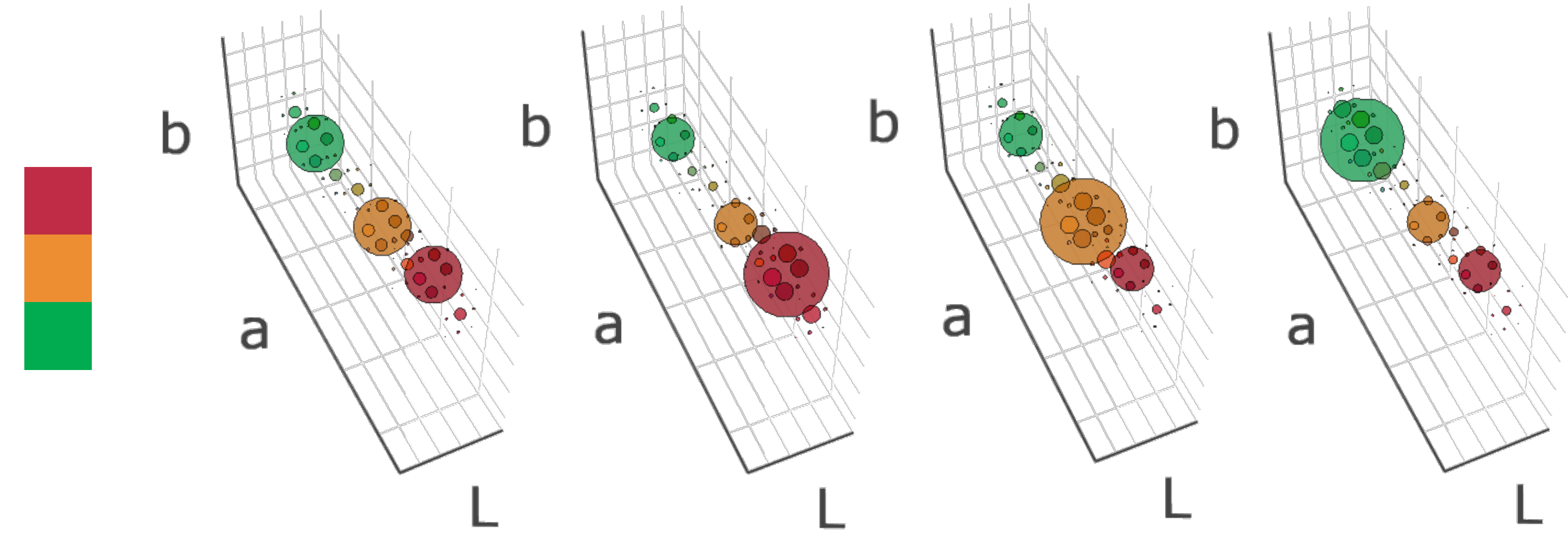}
\caption{We transform a color combination (left) to four histograms (right) according to the ratios 1:1:1, 2:1:1, 1:2:1, and 1:1:2. The sphere size indicates the bin size. Note that the histograms have been smoothed by Gaussian blur.}
\label{fig:histograms}
\end{figure}

\begin{figure*}
    \centering
    \includegraphics[width=\linewidth]{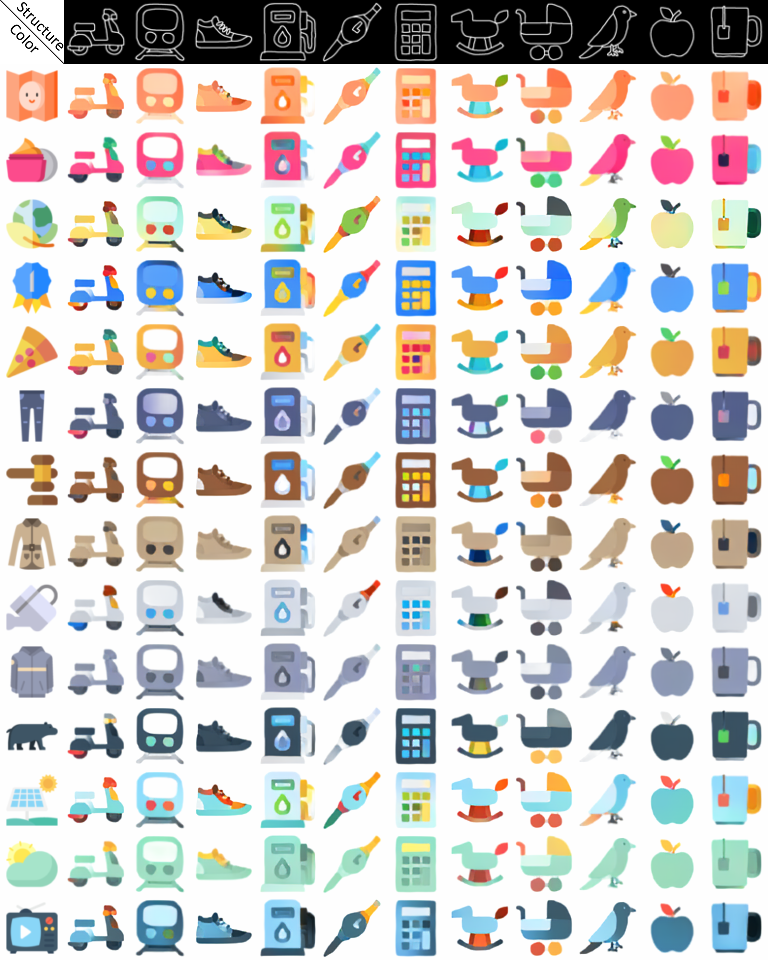}
    \caption{We applied our trained network to colorize icons in various styles. The first row shows contours, and the first column shows man-made icons that specify color conditions.}
    \label{fig:results}
\end{figure*}

\subsection{Training Details}
We trained the dual conditional GAN by using the Adam optimizer \cite{kingma2014adam} on a single NVIDIA GeForce 1080Ti. The learning rate was set to $10^{-4}$; the hyper parameters were initialized by using Xavier initialization \cite{glorot2010understanding}; and the batch size was set to 64. In each epoch, the generator $G$, the color discriminator $D_c$, and the contour discriminator $D_s$ were updated sequentially based on the stochastic gradient of $G^*$. We repeated the process 1,000 epochs until the loss was unable to decrease and the system started overfitting. It is worth noting that we did not use latent noise in the network.

\subsection{Semantic Style Labels}
Since applying a referenced icon to specify the color condition is not intuitive, we let users simply select a style label when using our system to create icons. Specifically, we consider the color psychology theory \cite{Kobayashi:1992:color} and define the style of a man-made icon according to its color combination. In our current implementation, each color combination contains three major colors; and each combination refers to a style. For example, an icon that contains \#C02C46, \#ED8E32, and \#01AC50 can be considered \emph{casual}, while an icon that contains \#94DEE2, \#FFFFFF, and \#91D2F1 can be considered \emph{clear}. Figure~\ref{fig:colorCombanations.pdf} shows the illustration. Accordingly, when the users select a style label, our system randomly selects a number of man-made icons that fulfill the style, and then feeds the icons to the network as color conditions. We recommend readers to watch our accompanying video for this intuitive user interface, as user interactions are difficult to be visualized in still images.

To determine whether an icon image $p$ matches a color combination (style) $q$, we compute their 3D Lab color histograms (8 $\times$ 8 $\times$ 8) for comparison. When determining the histogram of an icon $H_p$, white background pixels are not considered. In addition, because colors in adjacent bins can be visually similar, but are considered different, we apply 3D Gaussian blur to the histogram to reduce the difference between perception and statistics. Then, each bin is normalized by the total pixel number. Different to icon images, the definition of color combinations is rough. We generate four histograms $H_{q1}$ - $H_{q4}$ for each color combination based on the ratios (1:1:1, 2:1:1, 1:2:1, and 1:1:2) of three major colors, as illustrated in Figure~\ref{fig:histograms}. Finally, the icon is labelled as style $q$ if its color histogram $H_p$ is similar to either $H_{q1}$ - $H_{q4}$.

\begin{figure}
    \centering
    \includegraphics[width = \columnwidth]{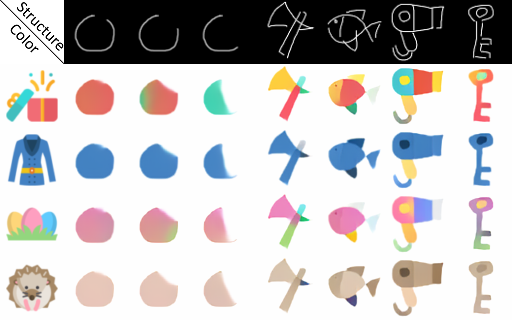}
    \caption{Our system can colorize icons that are conditioned by open contours without causing color leaking problems.}
    \label{fig:wipeout}
\end{figure}

\begin{figure}
    \centering
    \includegraphics[width = \columnwidth]{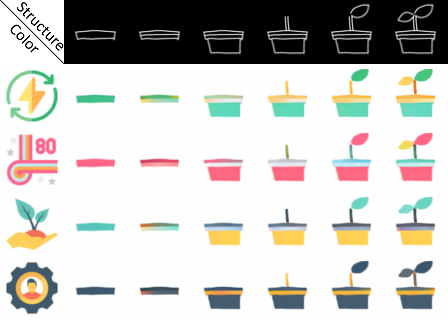}
    \caption{Step by step results. Our system generates results interactively while the input contour is incrementally enhanced. Each row shows the results based on the referenced icon image on the left. Notice that the colors were not changed considerably in subsequent steps once the referenced icon was selected.}
    \label{fig:step_by_step}
\end{figure}

\section{Results and Evaluations}
Several contour images, which contain straight and curved lines, and small and large open areas, were tested on our system. As can be seen in Figure \ref{fig:results}, these machine generated results were similar to man-made icons. For example, the results were in flat colors; only foreground objects were colorized; and most of the noticeable color boundaries could match the specified contours. The generated results are well elaborated and look like carefully designed icons. Interestingly, the results generated by considering the same reference icon were very similar in color style. This property is helpful to designers if they need to colorize a set of icons in a particular style. Considering that previous colorization methods may suffer from color leaking artifacts, we tested our system on several open contours. Figure \ref{fig:wipeout} shows that the system is leak-proofing. Users are not expected to draw contours carefully when using it.

The trained network can colorize icons in real-time by leveraging the GPU (graphics processing unit) resources. Hence, in our implementation, we feed the contour image and the referenced icon image into the network whenever a stroke is updated and then display the generated result immediately. Figure~\ref{fig:step_by_step} shows the results for the input contours that are incrementally enhanced.

\begin{figure}
    \centering
    \includegraphics[width = \columnwidth]{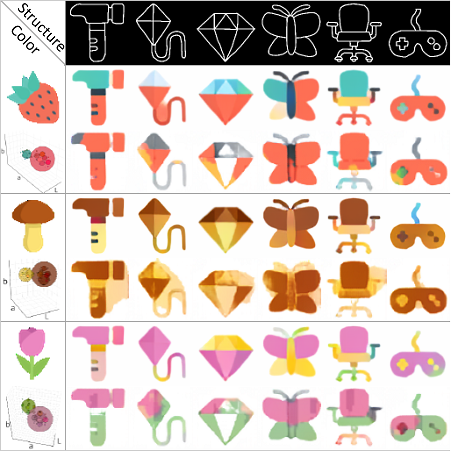}
    \caption{We applied different color conditions (i.e., image and histogram) to guide the network and compared their results.}
    \label{fig:spatial feature}
\end{figure}

\begin{figure*}
    \centering
    \includegraphics[width = \linewidth]{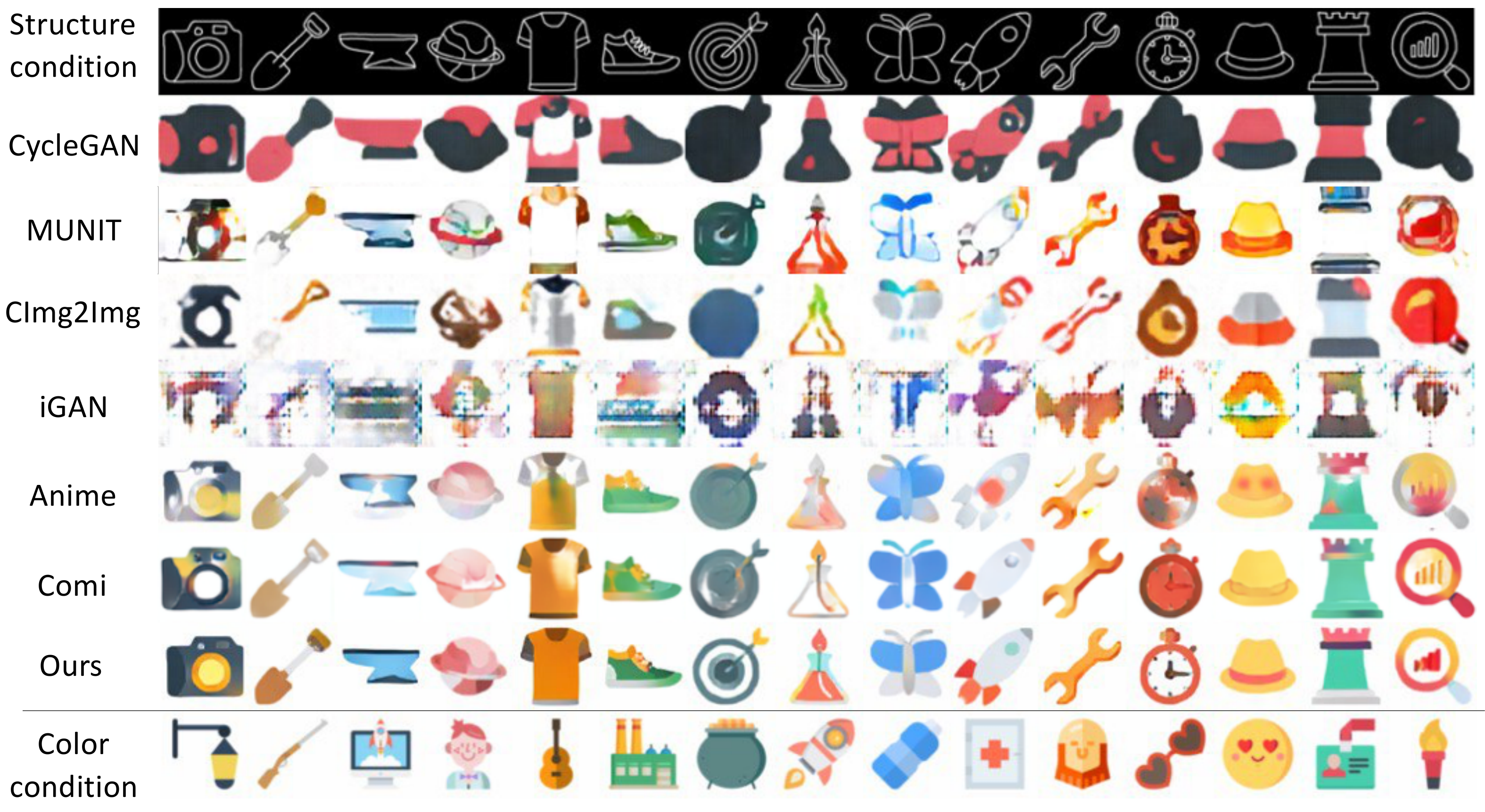}
    \caption{We compared our system to the current state-of-the-art techniques. From top to bottom rows are the contour images, the icons generated by CycleGAN \cite{Zhu+:2017:UnpairedImgImg}, MUNIT \cite{huang2018multimodal}, CImg2Img \cite{Lin+:2018:ConditionalIT}, iGAN \cite{zhu2016generative}, Anime \cite{zhang2017style}, Comi \cite{furusawa2017comicolorization}, our system, and the icons referenced by Anime and our system.}
    \label{fig:comparison}
\end{figure*}

\subsection{Representations of a Color Condition}
Several ways exist to represent the color condition of an icon. In addition to an image, we input a 3D Lab color histogram to the network and compared the results generated by these two different representations. Specifically, we duplicated the $8 \times 8 \times 8$ histogram to form an $8 \times 8 \times 256$ feature map. Therefore, on the generator part, the two feature maps determined from the contour image and the histogram can be concatenated. Regarding the discriminator part, we encoded the generated icon from $64 \times 64 \times 3$ to $8 \times 8 \times 256$ by three 2D convolutions so as to concatenate with the feature map of a color histogram.


Figure~\ref{fig:spatial feature} shows that both images and histograms are able to condition the color of the generated icons. However, the network can learn additional features if the color condition is represented by images. For example, a (nearly) closed region is in the same color, whereas adjacent regions are in different colors. In addition, we observed that the network adopts different strategies if the color and the structure conditions conflict with each other, which may occur when complexities of the contour image and the referenced icon image are considerably different. As shown in Figure~\ref{fig:spatial feature}, if the color condition is represented by a histogram, the generated icons may contain unnecessary edges/structures; if the condition is represented by an image, the generated and the conditioning icons may have different dominant colors.

\begin{figure}
    \centering
    \includegraphics[width = \columnwidth]{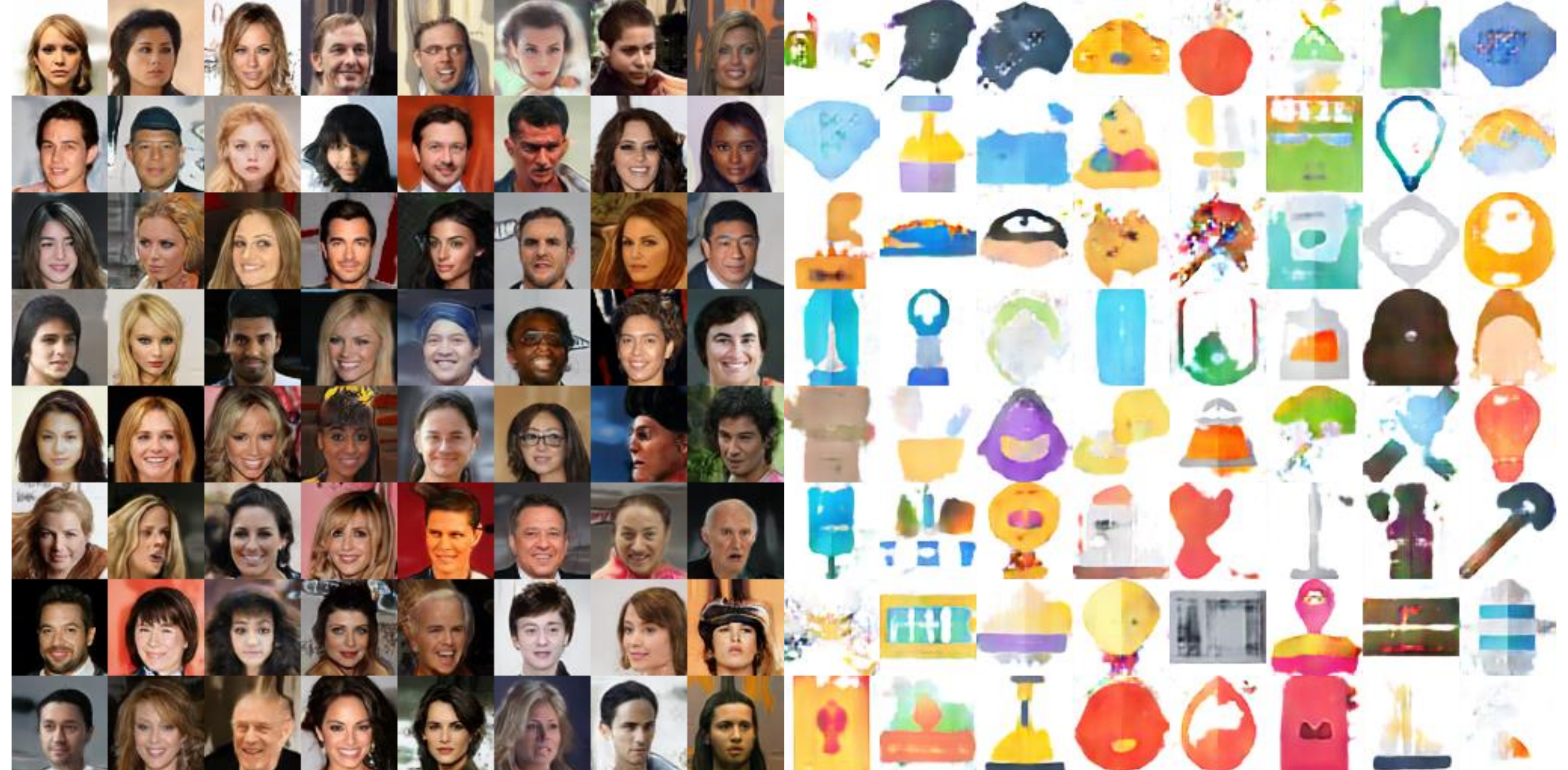}
    \caption{We trained a SNGAN \protect{\cite{miyato2018spectral}} on celeba and our collected icon datasets, respectively. It fails to generate icons (right) due to complex structures and styles.}
    \label{fig:SAGAN}
\end{figure}

\subsection{Comparison to State-of-the-Art Techniques}
We compared our method to the state-of-the-art techniques to evaluate its effectiveness. All the methods were trained on our collected icon dataset for the comparison. 

\textbf{iGAN. }
Interactive GAN (iGAN) \cite{zhu2016generative} can produce samples that best satisfy user edits in real-time. The system is based on DCGAN \cite{radford2015unsupervised:DCGAN}, which optimizes the latent vector to generate results specified by the color and the shape of brush strokes. As can be seen in Figure \ref{fig:comparison}, the results of iGAN are not satisfactory because low level features are noisy. In other words, although the generated icons to some extent fulfill the structure conditions, they are unsatisfactory.

\textbf{Domain Transfer Methods. }
CycleGAN \cite{Zhu+:2017:UnpairedImgImg}, CImg2Img \cite{Lin+:2018:ConditionalIT}, and MUNIT \cite{huang2018multimodal} are well known methods that can transform images from one domain to another. Thus, we were curious whether they were able to transform images in the contour domain to images in the color icon domain as well. The codes of these methods were obtained from the authors. As can be seen in Figure~\ref{fig:comparison}, given contour images, the results generated by their networks to some extent contain the features of icons, such as flat colors and simple lines. However, the results are not anticipated because their structures differ from the given contours. We suspect the reasons as follows. Given two domains $X$ and $Y$, the goal of domain transfer is to transform samples in $X$ to samples in $Y$, denoted as $Y'$, and then back to samples in $X$, denoted as $X'$. Another pass is from $Y$ to $X$ and then back to $Y$. The relations between $X$ and $Y$ are learned by the network itself. Normally, since the network has to transform samples in $Y'$ back to samples in $X$, the samples in $Y'$ must contain some features in $X$ to facilitate transformation. However, this does not mean that all features in $Y'$ are all related to $X$. A part of them is not used. For example, in the icons generated by CycleGAN, CImg2Img, and MUNIT, although a part of the edges can be matched to the contour images, a part of them cannot. In general, the redundant features make the results deviate from our expectation. If the domain $Y$ is narrow, such as a face dataset, the problem is not serious because the redundant features is suppressed by the discriminator. However, if the domain $Y$ is wide, such as the icon dataset, the redundant features become noticeable. That a GAN can generate face images from a random latent vector, but may fail to generate icons (Figure \ref{fig:SAGAN}), supports this assertion. Notice that the faces are recognizable, although there are many redundant features (artifacts).



\textbf{Manga/Cartoon Colorization. }
The goal of our system is similar to Manga colorization. Both of them attempt to colorize contour images. Therefore, we compared our system to a famous online software called style2paints\footnote{https://github.com/lllyasviel/style2paints}, which was an improved version of \cite{zhang2017style}. However, because the details of this online software was not disclosed and the network was trained on Manga images, the comparison may not be fair. We further implemented the original network \cite{zhang2017style} and trained the network on our collected icon dataset. Figure \ref{fig:comparison} shows the results. As can be seen, the interior structures of the icons generated by the method (Anime) are barely matched with the input structures. The network of \cite{zhang2017style} does not function well because the discriminator is trained to determine man-made and machine-generated icons. The domain is too wide and the task is too difficult. In addition, the training strategy makes the generator overfit easily. During the training phase, the input contour and style images are very similar in structure. However, during the testing phase, the two images are different.

We also compared our system to Comicolorization \cite{furusawa2017comicolorization}. The codes were obtained from the authors. In the beginning, we failed to train the network by following their procedure because the discriminator cannot recognize icons. Hence, we reduced the weight of the adversarial loss to 0.05 and then succeeded. In other words, the network was trained mostly based on the reconstruction loss. Since the supervised learning tends to fit the whole data distribution, the generated icons often contain blurring artifacts. The structures of their generated icons (Comi) are not as sharp/clear as the structures of ours (Figure \ref{fig:comparison}).

\begin{figure}
    \centering
    \includegraphics[width = \columnwidth]{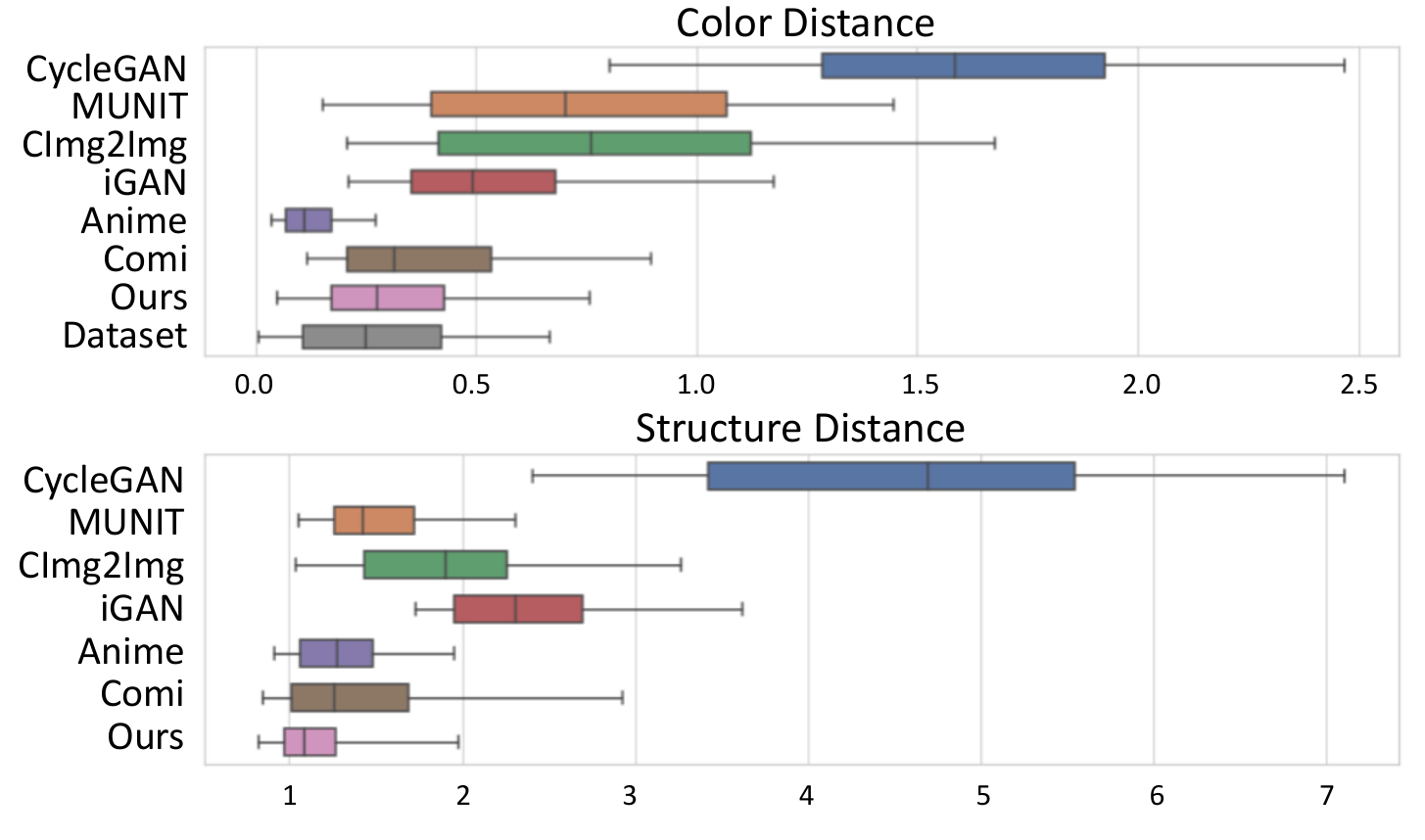}
    \caption{Fulfillment of color and structure conditions among the previous and our methods. The five-number summary (from left to right on the box and whisker plots) consists of the $5^{th}$, $25^{th}$, $50^{th}$, $75^{th}$, and $95^{th}$ percentiles.}
    \label{fig:quantitative}
\end{figure}

\subsection{Objective Evaluations}
In addition to visual comparison, we quantitatively evaluated the generated results. Color distances and structure distances between the generated icons and the conditions were computed. To measure the fulfillment of color condition, we computed the Jensen-Shannon divergence of 3D Lab color histograms for the evaluation. To measure the fulfillment of structure condition, we first applied the Canny edge detection method to the generated icon image. Afterward, bi-directional search of the closest edge pixels is adopted to compute the distance between the generated and the conditioned contours. Specifically, for each edge pixel $p$ in one image, we searched the edge pixel $q$ in the other that is closest to $p$, and computed the mean distance $D_{pq}$. The two images were then switched for estimating the mean distance $D_{qp}$. In other words, we measured the fulfillment of structure condition by ${1 \over 2} (D_{pq} + D_{qp})$.


The box and whisker plots in Figure \ref{fig:quantitative} show the evaluation results. Clearly, CycleGAN, MUNIT, CImg2Img and iGAN could not fulfill the color condition because of the monotonic or noisy colors. It is worth noting that the icons generated by Anime were the most similar to the referenced icons in colors. By comparing the statistic and the results shown in Figure \ref{fig:comparison}, we found that Anime tends to copy colors from one image to another when colorizing an icon. However, this strategy is inappropriate because the contour image and the referenced icon were different in structure. The number and the sizes of contours cause the conflict. To verify this inference, we randomly picked two man-made icons that were classified in the same cluster (Section \ref{secton:colorcondition}) and computed their mean color distance. The statistic shows that man-made and our colorized icons were well matched. In contrast to the color distance, the shorter contour distance is the better because ideally we expect the generated icon and the contour image to be the same in structure. Again, the distances show that icons generated by CycleGAN, MUNIT, CImg2Img, and iGAN were dissimilar to the specified contours. Anime and Comi could fulfill the overall structure condition. However, our method did a better job in colorizing details.

\subsection{Subjective Evaluations}
We conducted a user study with 92 participants to evaluate the results generated by CyleGAN, MUNIT, CImg2Img, iGAN, Anime, Comi, and ours. Specifically, we created a questionnaire and posted it in the Internet for anonymous participants to answer. They were asked to compare and to rate icons colorized by different methods. The best to the worst icons were rated by 5 to 1, respectively. To achieve a fair comparison, the icons that were conditioned by the same color and the same structure were listed in a page. In addition, the order of the methods was randomly assigned to prevent bias.

The mean scores rated by the participants to our method, Comi, Anime, MUNIT, CycleGAN, CImg2Img, and iGAN were 3.65 (SD = 1.12), 3.27 (SD = 1.20), 2.67 (SD = 1.14), 2.07 (SD = 1.02), 1.66 (SD = 0.80), 1.57 (SD = 0.72), and 1.25 (SD = 0.64), respectively. As indicated, our method was rated the best. To understand whether the result had statistical significance, we ran a one-way ANOVA to analyze/compare the scores of the previous methods and ours. The results confirmed that our system was qualitatively better than the other methods ($p < 0.01$ for all of the comparisons).

\begin{figure}
    \centering
    \includegraphics[width = \columnwidth]{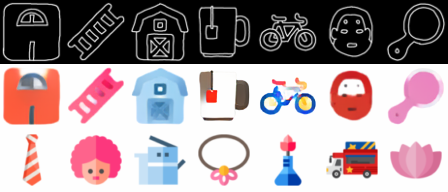}
    \caption{Several failure examples generated by our system. From top to bottom are the contours, results, and the reference icons.}
    \label{fig:limitation}
\end{figure}

\subsection{Limitations}
Generating results that are always satisfactory in semantics is difficult. Figure \ref{fig:limitation} shows several failure examples. Specifically, some icons, such as an apple, a lion, or a tree, have their own colors. Our system is likely to generate results that do not match the target semantics (e.g., colorizing an apple with blue) because it only considers whether the generated and the reference icons have the same color style. We also find that several external regions, such as the holes formed by the steps of a ladder and the bicycle frame, are mis-colorized. In addition, the structure and color conditions can conflict when the complexities of a contour image and a referenced icon are different. If a simple contour and a diverse-color icon are fed into the generator, the generated result may contain additional boundaries or gradient colors that should not appear in icons. Regarding the combination of complex contour and monotonous color, the generated results are of two types: 1) they can be as monotonous as the reference icon; or 2) they contain colors that are not in the reference icon.

\section{Conclusions and Future Works}
We have presented an interactive system to help designers create icons. This is a system that allows both humans and machines to cooperate and explore creative designs. Specifically, designers draw contours to specify the structure of an icon; then the system colorizes the contours according to the color conditions. This goal cannot be achieved by previous methods due to various structures and styles of man-made icons. Training a discriminator to recognize a man-made or a machine-generated icon is very challenging. Therefore, we divide the icon recognition task into two sub-tasks and apply a dual conditional GAN to solve the problem. Specifically, the two discriminators determine whether the structure and the color of paired images are well matched, respectively. While the generator successfully cheats these two discriminators, it can generate icons demanded by users.

The color condition of our current system is represented by an image. To improve usability, we let users specify a style label when using our system to create icons. Man-made icons that fulfill the label will be randomly selected and then fed into the network as color condition. In other words, the network has no idea about styles. Therefore, in future, we plan to train a network that can take semantic labels as input when colorizing icons. Considering that style labels are implicit, we believe that the strategy also has the potential to solve the conflict of structure and color conditions.

\section*{Acknowledgements}
We thank anonymous reviewers for their insightful comments and suggestions. We are also grateful to Prof. Chun-Cheng Hsu for the valuable discussions, and all the participants who joined the user study. This work is partially supported by the Ministry of Science and Technology, Taiwan, under Grant No. 105-2221-E-009 -135 -MY3 and 107-2221-E-009 -131 -MY3.

\bibliographystyle{ACM-Reference-Format}
\balance
\bibliography{acmart}

\end{document}